\documentclass{article}
\usepackage{fullpage,enumitem,amsmath,amssymb,graphicx,qtree,parskip,multicol}
\begin{document}

\title{Autonomous Haiku Generation}
\author{Rui Aguiar and Kevin Liao \\
(raguiar2@stanford.edu, liaok@stanford.edu)}
\date {\today}
\maketitle
\section*{Introduction}
Artificial Intelligence is an excellent tool to improve efficiency and lower cost in many quantitative real world applications, but what if the task is not easily defined? What if the task is generating creativity? Poetry is a creative endeavor that is highly difficult to both grasp and achieve with any level of competence. As Rita Dove, a famous American poet and author states, "Poetry is language at its most distilled and most powerful."$^{[1]}$ Taking Dove’s quote as an inspiration, our task was to generate high quality haikus using artificial intelligence and deep learning.

\section*{Task definition}
Our task can be summarized as "given the first line of a haiku poem, autonomously generate the next 2 lines." For example, if the first line of a poem was 

barren trees

\noindent We might try and construct the rest of the poem to be coherent given the first line, outputting something similar to the actual poem, along the lines of:

barren trees \\
\indent even the tiniest twig \\
\indent embraced by the mist

\noindent Our input was the first line of a haiku poem from a test set, and our output was the next two lines of the poem. We attempted to generate the poems in roughly the 5-7-5 format that traditional haikus tend to follow, though there were some problems with this format that we discussed in our error analysis section.

\noindent Initially, we decided to generate tanka poems due to the structured nature of the poem's form. However, due to the lack of proper  datasets, as well as the longer nature of the poem that made the search states large and difficult to generate contextual poems, we decided to switch to haiku poems. 

\noindent After creating an AI to generate these poems, we evaluated them by asking 8 people to score our poems, alongside some baseline and oracle poems in a single blind trial to evaluate the quality of our poems and their performance relative to oracle and baseline. The participants in the evaluation process were given two questions:

\indent 1. ``On a scale of 1-10, how good was this poem" \\
\indent 2. ``On a scale of 1-10, how likely was it that this poem is written by a human?" 

\noindent This evaluation is further discussed in the results part of our paper. We segmented our approach down into two main parts: A beam search coupled with a machine learning model to generate haikus, and a recurrent neural network on both a character-based and word-based level.

\section*{Infrastructure} 
Before we could begin generating poems, we had to first collect a dataset and perform the needed infrastructure work to get the data in the correct format for training and text generation. We initially scraped and parsed several hundred poems from http://www.bartleby.com/ and inserted them into a text file using a python script as our database. As the scope of our project shifted and we required more data to train our neural networks, we quickly realized that this dataset would not be enough, and so we used a large haiku dataset found in a CSV on a Github repository.$^{[2]}$ The dataset contained over 8000 haikus scraped from popular poem websites, and served as enough data to train our neural networks and other models.

\noindent Both throughout the project and as we moved to our new dataset, we ran into numerous text-parsing problems, from ACII-Unicode conversion to the line break character differing between lines, ranging from tab to the EOL character. This required some infrastructure/cleanup work on our dataset to resolve. We also did significant file I/O work, especially in our beam search iteration of results to manipulate the first lines into a separate file so we could sample randomly from training and test sets to generate and validate our poems. We also had to implement a lot of file I/O when working with our neural network models/saved weights, as we trained and saved our model, and then loaded it into multiple other files.   

\noindent Beyond the file I/O and web scraping, we used numerous third party tools in order to achieve the best results for our poetry. We also created several scripts that allow users to interface with our poem generation tools, and generate poems for themselves. The entire code base can be found on github.$^{[3]}$

\noindent In our baseline approach, we used several utilities from our old CS221 assignments, notably Reconstruction's unigram and bigram costs. We modified these utilities to fit our data format and file I/O, but the inspiration for the cost functions came from these assignments. Additionally, we used python's nltk and curses library to get the number of syllables in the words we were generating. 

\noindent For our Beam Search iteration, we used numerous third party tools in our infrastructure. Firstly, we used a modified iteration of the Beam Search problem from a past assignment with Uniform Cost Search. To generate and log about our word2vec model, we used python's gensim and logging libraries. We also used matplotlib to plot the progress of our machine learning model across training and test data. 

\noindent When we switched to using a Recurrent Neural Network to generate poems, we had to make extensive use of third party libraries to help with our poetry generation. We used keras, a python deep learning library built on top of TensorFlow to generate and train our RNN. We also switched to Pyphen, a hyphenation library with a syllable count method for our syllable counts, because it had a more reliable counter when the word we generated was not in the English dictionary or was misspelled, something that happened with relative frequency, especially when we did character-level generation. 

\noindent Overall, our heaviest infrastructure work came from creating web scraping tools to get poems for our dataset, creating scripts so users could interact with our model, and using several third party libraries for clean and effective code.

\section*{Approaches}
\subsection*{Baseline Poems}
For our baseline algorithm, we entered the first line of a haiku and used a greedy algorithm with a bigram cost to select the next word, only stopping once we had gone over the maximum number of syllables allowed in that particular line. This method sometimes meant we sacrificed the form of the poem slightly to reduce the bigram cost. However, we still produced structurally sound, though a bit nonsensical, poems. 

\subsection*{Oracle Poems}
Our oracle is human generated haikus that get high scores on quality and probability of being written by a human. To get our oracle poems, we used published haiku poems which we randomly selected from our dataset. There is a significant gap in quality between the baseline and oracle. While the baseline produced poems that are structurally correct, the oracle is much more coherent, and uses related concepts to the theme of the poem. Additionally, the oracle's poems are much more likely to form a coherent message than a poem generated with simple bigram cost. The challenge of this project comes from closing that coherence gap, as well creating poetry that pertains to the subject matter and makes sense, given a first line of input.

\subsection*{Beam Search}
For our initial iterations of a solution, we decided to generate our poems using a Uniform Cost Search, which we later switched to a Beam Search approach. The reason that we used Beam Search was because we thought that this would be a way to generate the poems that would have whatever we defined to be the lowest cost, and the natural structure of the haiku lent itself to a Beam Search. Initially we began with a Uniform Cost Search approach, but after pivoting to haikus and using a larger dataset, we realized that Uniform Cost Search would take too long to generate due to the massive amount of words in the new larger training set. Instead, we decided to modify it into a Beam Search with the best 20 words as the branching factor.

\noindent For our cost function, we used machine learning to train a model that would allow us to better predict the next word in a poem. To do this, we used word2vec, a software that lets you generate a vector representation of a word. Word2vec is a 2-layer neural net that produces a vector corresponding to each word in the corpus. We used this to find the similarity of the two words by taking the dot product of the vector representation of these words. We trained a word2vec model for each word in our dataset; then generated examples of the form:

(beginning of line excluding last word, similarity of (penultimate word, last word)) 

\noindent where the similarity was how close word2vec predicted the two words to be in meaning. Essentially, our final model attempted to predict the similarity of the word following a previous sequence of words. 

\noindent While training this model, some of our main features included the length of the current line (in number of words), the unigram cost of the previous word, bigram costs of previous words used in the sentence, and weights corresponding to each word used in the poem. Our loss function was the squared loss function used in linear regression. After numerous iterations, our model was able to more accurately predict the similarity of the next word in the poem. The graph showing the average difference in similarity of words predicted after 1000 iterations on our validation data is displayed below (Figure 1): 

\centerline{\includegraphics[width=7cm]{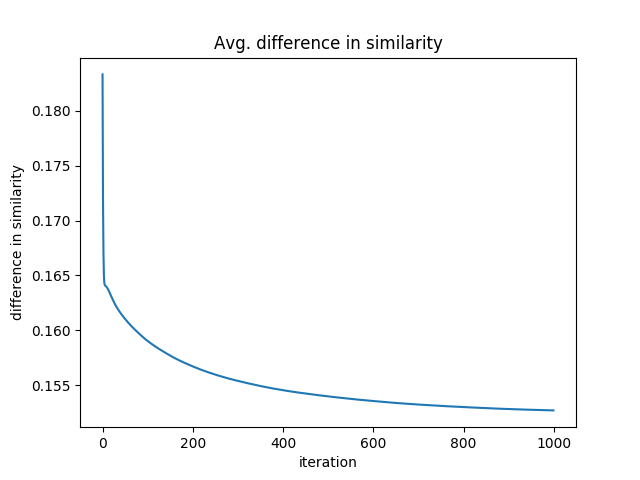}}
\centerline{Figure 1.}

\noindent With each iteration, the model becomes better at predicting the similarity between words, and the difference in similarity goes down.

\noindent After training this cost function, the costs in our Beam search were the absolute value of the difference between the similarity of the candidate word and what our model predicted the similarity of the next word to be. Our Beam search can be described mathematically as 


\noindent Each state is (previous\_word, current\_line, num\_syllables)\\ 

\noindent $S_{start}$ = (firstline[-1], "",0) \\
$Actions(s)$ = [k\_most\_similar\_words(s.previous\_word,k=20)]\\
$Cost(s,a)$ = $\mid model.similarity(previous\_word, a) - predicted\_similarity(previous\_word) \mid$ \\
$Succ(s,a)$ = (new\_word, s.current\_line +" "+new\_word, s.num\_syllables + num\_syllables(new\_word))\\
$IsEnd(s)$ = (syllables == self.max\_syllables)\\

\noindent The poems generated from Beam Search will be discussed in the "Results" section, but to further improve our poems' quality with respect to the oracle, we have to integrate more context when making the word choice. Hence, we decided to continue our project with the Recurrent Neural Network approach.

\subsection*{Recurrent Neural Network}
For the Recurrent Neural Network approach (RNN), specifically using Long Short Term Memory cells (LSTM's) that captured the context of prior predictions and text to make the next prediction, we used a stacked model, as shown in the diagram to the right below:

\centerline{
\includegraphics[width=7cm]{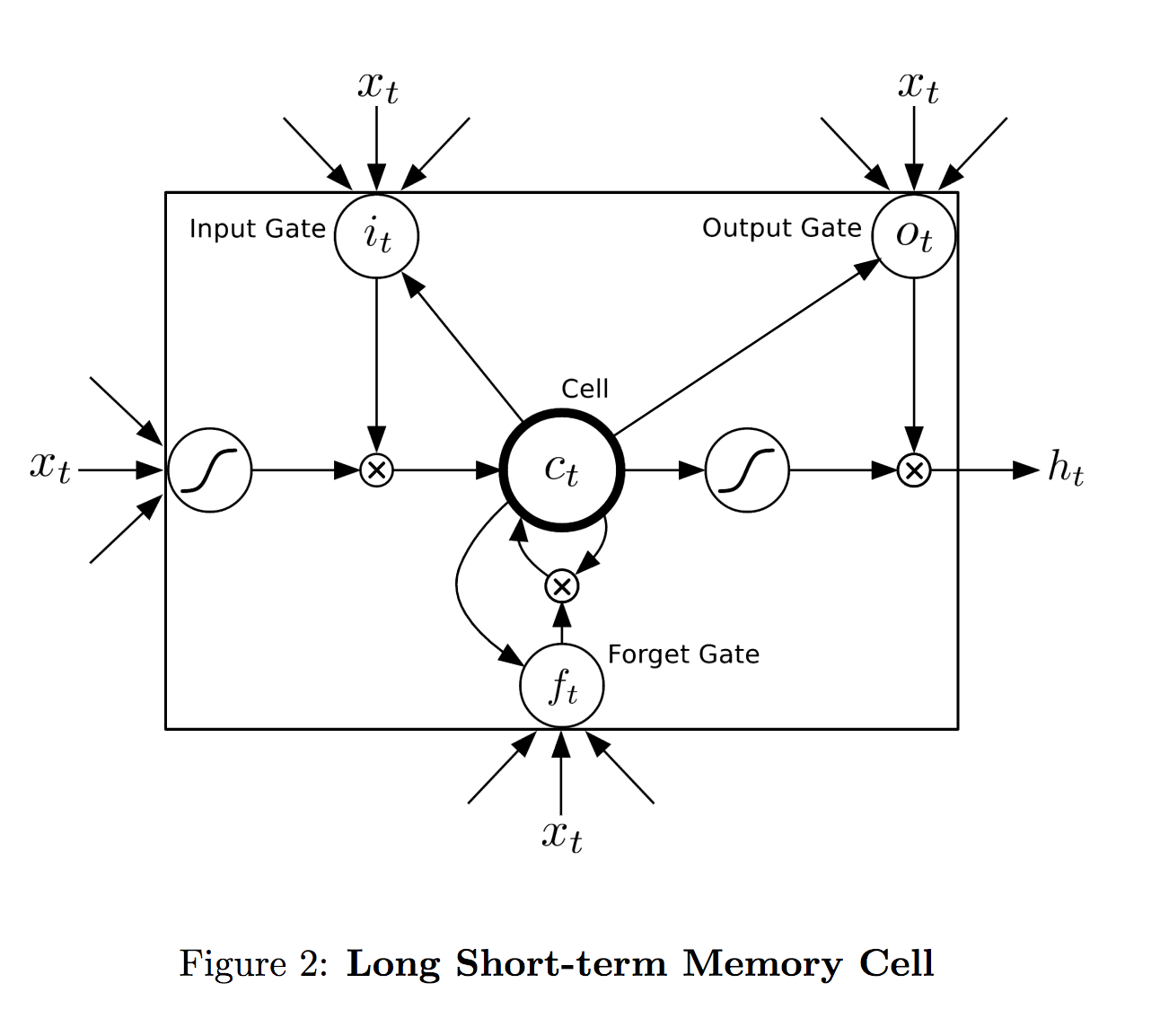}
\includegraphics[width=6.5cm]{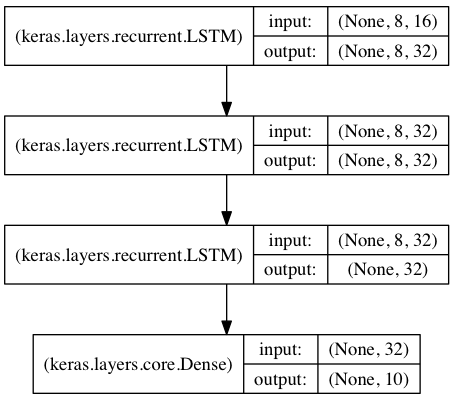}
}

\centerline{Figure 2.$^{[4]}$ \hspace{14em} Figure 3.$^{[5]}$}

\noindent In our first iteration of our RNN, we trained the algorithm with our dataset by splitting them up into pairs of (window\_of\_characters, next\_character). We train our RNN with these pairs to get probability distributions for the next\_character. For poem generation, we took in the sliding window of 10 characters (initially containing the first line that the user inputs) and from that we obtain a probability distribution of the next character to be appended to the window of characters. We then added some mathematical noise to our probabilities, and selected the character with the highest probability after the noise was added.$^{[6]}$ Then we continue until we choose an end-of-line character, where we subsequently start the next line. Once the third line is finished, we have our new haiku. 

\noindent While our character level model was actually able to generate coherent words and poems the majority of the time, we were having issues with nonsensical words being generated, and decided to switch to a word-based model. Our word based model worked similarly to our character based model, with a few key differences. Firstly, instead of generating the poems one character at a time, we began generating them one word at a time. Additionally, instead of having a sliding window of previous characters, we were able to take in a sliding window of previous words and use those as our data points to predict the next word in the poem. Empirically, we were able to get better results with this model, so we decided to stick with word-based generation for our final model. 

\noindent One of the main challenges in both of our RNN models was correctly tuning the parameters and hyperparameters to minimize validation loss while being careful to avoid over fitting the model. This took a lot of time and we used several techniques to achieve the optimum loss subject to the over fitting constraint. 

\noindent First, we experimented the number of epochs we trained the data on, eventually settling on 25 epochs, which we found generated the model with the smallest validation loss without beginning to overfit. We also optimized the learning rate to be as large as possible, so we could make the largest steps while still converging to a decreasing loss. We also worked with the batch size to reduce overfitting, as when our batch size was too small, the model began to overfit after less than half the iterations. 

\noindent To further reduce the loss, we also experimented with the layers we used in our RNN. We had a layer of LSTM cells, and after experimenting with several values, decided to use 512 LSTM cells. We also added a dense layer and a dropout layer, the latter of which served the purpose of excluding some weight updates to reduce over fitting. 

\begin{center}
\includegraphics[width=8cm,scale=0.3]{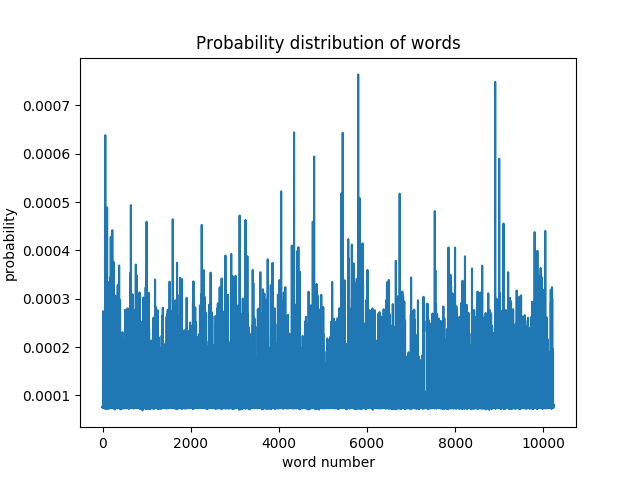} 
\includegraphics[width=8cm,scale=0.3]{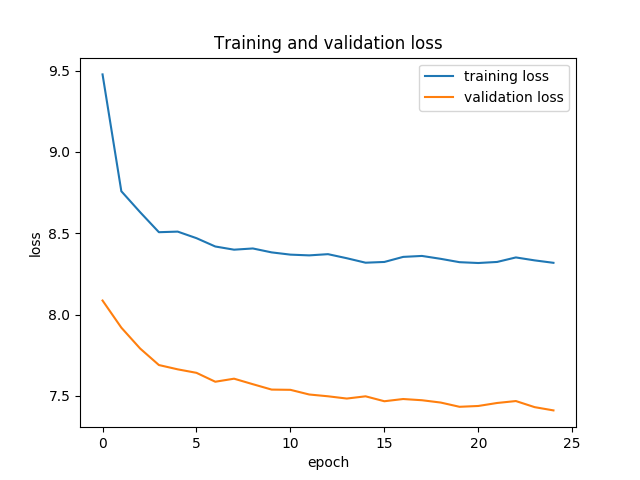} \\
Figure 4. \hspace{18em} Figure 5.
\end{center}

\noindent In summary, our approach took several iterations, each of increasing complexity. Our baseline was simply generating text with a greedy algorithm, which we then expanded into a Beam Search/ machine learning approach. Finally, we switched to using recurrent neural networks, which took extensive parameter tuning to optimize. As we changed our model, our results improved drastically across each iteration.

\section*{Results} 
For our results, we asked 8 people to evaluate our poems based on the two questions:

\indent 1. ``On a scale of 1-10, how good was this poem" \\
\indent 2. ``On a scale of 1-10, how likely was it that this poem is written by a human?" 

\noindent In this section, we will discuss some examples of the poems generated, as well as how the poems at each iteration of our approach compared to the prior iterations. For each graph in this section, we took the average of all the scores on each of the two poems we asked our evaluators to score in each category. 

\noindent One of the first things we did for poem comparison was select two oracle poems from our database of haikus at random to compare our results to it.  Our oracle poems were:

\begin{multicols}{2}
\noindent
withering leaves\\
the lawyer to write his will\\
rings the doorbell
\\
a hawk's eye?\\
autumn gusts rattle\\
the beech leaves
\end{multicols}

\noindent Note that these poems do not quite follow the 5-7-5 format. We found that most of the poems in our dataset did not, and we will discuss this in the error analysis section. 

\noindent As discussed before, our baseline poems were generated using a greedy algorithm, where we looked at the previous word in the poem, sometimes the last word from the previous line, and selected the word with the lowest bigram cost as the next word in the poem. We ended the line once we got to or exceeded the syllable count. Our justification for exceeding the syllable count was that it would lead to more coherent poems if our bigram cost was lower, and our dataset did not follow the 5-7-5 format very strictly. Here are some of the examples of poems generated using the greedy algorithm: 

\begin{multicols}{2}
\noindent
and so i agree\\
june junk glint stars... clicks\\
branch a bell tolls changes quails
\\
an eighth grader's voice\\
or six cub other\\
hip roars visitor's lovers...
\end{multicols}

\noindent These poems are somewhat nonsensical, and we were able to improve on them as we increased the complexity of our model. 

\noindent Our beam search/ machine learning model produced poetry with slightly more sense and structure. Some of the poems produced by this iteration included:

\begin{multicols}{2}
\noindent
the billowing cloud \\
a dragonfly breeze web \\
turnip uprooted
\\
year of the dragon \\
paddocks rides overdue \\
warming iron chill
\end{multicols}

\noindent This model began to show some progress over the baseline, both empirically and in the evaluation metrics.  The Beam Search generated the word 'uprooted' after 'turnip', and used the words 'breeze' and 'cloud' in the same poem, likely because they were related to each other and used in a similar context in our training set. This is different from the baseline because our cost model takes into account the similarities of words from the training set. Other poems we generated also had similar words mentioned in the same poem. 

\noindent Our Beam Search poems scored much better than our baseline poems with human evaluators, but there was still a significant gap between our oracle poems and those created by the Beam Search.

\begin{center}
\includegraphics[width=8cm,scale=0.3]{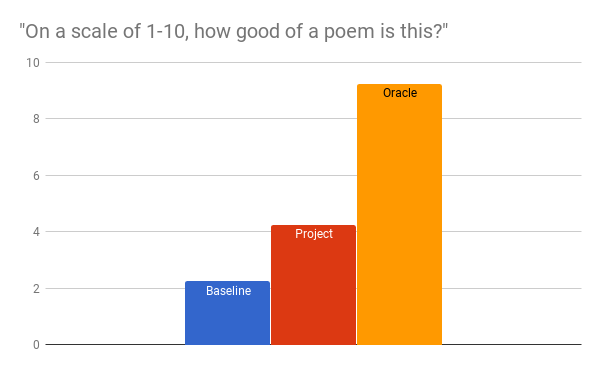}
\includegraphics[width=8cm,scale=0.3]{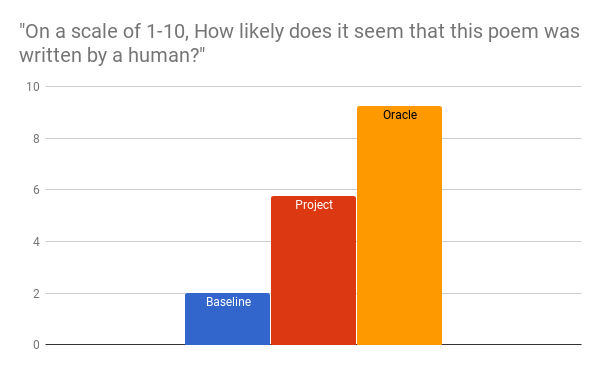}
\end{center}

\noindent Once we shifted to a RNN, more specifically the word-based model of our RNN, our poems scored much higher among reviewers than any previous iteration. Some examples of poems generated include:

\begin{multicols}{2}
\noindent
late afternoon a breeze sweeping \\
i pick down the grey beat of \\
our conversation

gliding hawk \\
moves all the day her stillness \\
the hill leaves its won
\end{multicols}

\noindent We asked our reviewers to evaluate these poems as well, and we then graphed how they scored against our oracle and Beam Search approaches. 

\begin{center}
\includegraphics[width=8cm,scale=0.3]{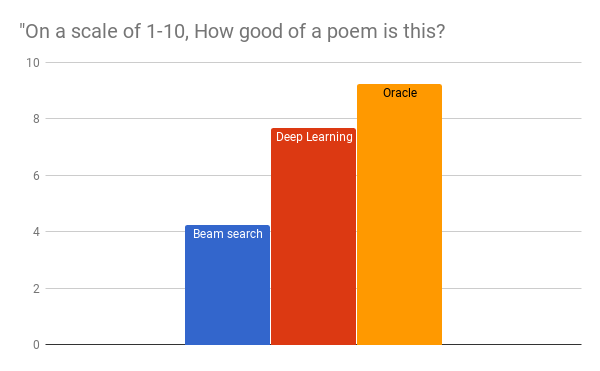}
\includegraphics[width=8cm,scale=0.3]{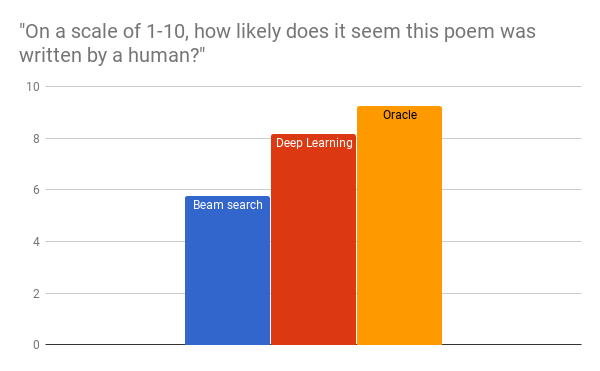}
\end{center}

\noindent Though there was still a gap between our oracle and deep learning poems, the ones generated using our RNN performed much better than our Beam Search poems in both parts of the human evaluation. 

\noindent Ultimately, as we increased the complexity of our model, our results improved. Though both our Beam Search and RNN scored better than the baseline, our RNN scored far better than our Beam Search. If we were to continue this project and modify parameters/experiment with new models, we could likely perform even closer to the oracle.  

\section*{Literature Review}
There have been several attempts at autonomous poetry generation, including several from previous CS221 projects. Some past iterations of projects that have done poetry generation include "Poetry Generation in Esperanto" by Angelica Previte and "Creating Sonnets from News Articles" by Chelsea Ann Hunter.  Additionally, there have been several attempts at autonomous poetry generation from outside of the context of CS221. One such attempt is "An evolutionary algorithm approach to poetry generation" by Hisar Maruli Manurung, which details natural language generation in the context of evolutionary algorithms.$^{[7]}$ There has been a lot of work using RNN's for text generation, including one article we based our initial model off of, titled "Using AI to generate lyrics.$^{[6]}$ Additionally, Google's project Magenta tries to create art and music using machine learning.$^{[8]}$

\noindent Though it is difficult to match the complexity and scope of industry projects or thesis papers, our approach does differentiate itself from past CS221 projects regarding poetry generation in that it uses recurrent neural networks for poetry generation, and uses the Word2Vec model to penalize word differences in our beam search approach.

\section*{Error Analysis}
Though our approach uses novel methods for poetry generation, it is not free of error. There are several sources of error in our project, the main issues stemming from dataset quality, incorrect syllable count in poem generation, and parameter optimization/model accuracy. 

\noindent One large source of error in our project was the quality of the dataset. We first scraped the majority of our poems from http://www.bartleby.com/ and used that as our dataset. However, as time progressed and we switched to haiku over tanka generation, we realized that we did not have nearly enough poems, and set out to find a large haiku dataset. Eventually, we came across one on a github repository that scraped haikus from popular poetry websites. While this provided us with an appropriate quantity of haikus, we soon realized that not all of the haikus were formatted properly. Often, what should have been a ' ' (space) character showed up as a '?' (question mark character). We suspect that this was an ASCII-Unicode conversion error when the web scraping occurred. This led to some poems that were nonsensical and used several words chained together by question marks as one word. Even after we filtered out these question marks, sometimes there were spaces interspersed randomly throughout poems, leading our model to classify some single characters (such as the character 's') as entire words, and creating poems that used this character as an entire word. We found the errors in the dataset too diverse and hard to pin down to correct with one hundred percent accuracy. Luckily, most of our poetry is not affected by this source of error. 

\noindent Another source of error related to the previous was the syllable count in our haikus. When doing word based models such as in our Beam Search and later iterations of our RNN, we found that many syllable counting functions provided by python libraries that we experimented with were not expansive enough to capture every single word in our poems, especially since the artists sometimes made up or hyphenated multiple words together in their poems. This led to our generated haikus occasionally not being of the 5-7-5 form, that traditional haikus are. However, as we investigated our dataset and haiku structure further, we realized that a large portion of the haikus in the dataset itself did not follow a 5-7-5 format, and so even in our character generation RNN format, we were modeling off of incorrect data and could not reliably produce 5-7-5 poems. Because our dataset of haikus itself did not follow this structure, and multiple library functions failed to produce reliable results, we were ultimately unable to consistently get poems in the 5-7-5 format. Additionally, from a poetry point of view, we realized that haikus in the modern day are not as strictly defined as compared to traditional haikus in Japanese where the poetic form originated. Some poets believe that there are more innovations developed when the traditional format is simply ignored.$^{[9]}$ Hence, haikus have developed a more flexible form in modern times, and we decided to go with this more unstructured definition of haikus for our project for the sake of both convenience and better poems.

\noindent One final source of error is variation in poem quality given our RNN. Often, our RNN produces poems that are relatively coherent. However, sometimes the poems generated by the RNN can still be somewhat meaningless. This may be because of variation in the training and test data, and different words appearing in the test data that the model cannot fit well on. 

\section*{Conclusion}
In summary, we found that for the most part, an RNN is the most effective  approach for text generation. While we had some slight variation in the quality of poems produced, we were able to generate coherent and meaningful poems at a much higher and more consistent level using the RNN than any other of our approaches. While our beam search approach outperformed the baseline, it was not able to match our RNN's performance. Additionally, though we produced a model to generate poems with decent quality, there could still be future work done for this project. We could model the generation of haikus as a constraint satisfaction problem to more closely match the 5-7-5 format. With a more complicated model, we may also be able to eliminate the inconsistencies in quality of our RNN-generated poems. While this project made significant progress on autonomous generation of poems, there is still a lot of room to improve and innovate. 

\section*{Codalab}
We created a Codalab worksheet for our project, which can be accessed at \\
https://worksheets.codalab.org/worksheets/0x7f7b9a8e7e544b19a78ef5fc7cc2ce8f/

\section*{Bibliography}
[1] Lannan Foundation. “Rita Dove.” Lannan Foundation, lannan.org/bios/rita-dove.

\noindent[2] Ballas, Sam. “PoetRNN.” Github, 1 Aug. 2015, github.com/sballas8/PoetRNN/tree/master/data.

\noindent[3] Aguiar, Rui. “haiku\_generation.” GitHub, 7 Dec. 2017, github.com/raguiar2/haiku\_generation.

\noindent[4] Stack Overflow. “How Does LSTM Cell Map to Layers?” Stack Overflow, Aug. 2017, stackoverflow.com/questions/45223467/how-does-lstm-cell-map-to-layers.

\noindent[5] Keras. “Getting Started with the Keras Sequential Model.” Guide to the Sequential Model - Keras Documentation, keras.io/getting-started/sequential-model-guide/.

\noindent[6] Liljeqvist, Ivan. “Using AI to Generate Lyrics – Ivan Liljeqvist – Medium.” Medium, Medium, 5 Dec. 2016, medium.com/@ivanliljeqvist/using-ai-to-generate-lyrics-5aba7950903.

\noindent[7] Manurung, Hisar Maruli. “An evolutionary algorithm approach
to poetry generation.” University of Edinburgh, Institute for Communicating and Collaborative Systems, 2013, www.inf.ed.ac.uk/publications/thesis/online/IP040022.pdf.

\noindent[8] Magenta. “Welcome to Magenta.” Magenta, 1 June 2016, magenta.tensorflow.org/welcome-to-magenta.

\noindent[9] North Carolina Haiku Society. “What's a Haiku.” North Carolina Haiku Society, nc-haiku.org/whats-a-haiku/.

\end{document}